\documentclass[journal,twoside,web]{ieeecolor}
\makeatletter
\def\@IEEEmembership{}
\makeatother
\providecommand{\refname}{References}
\usepackage{generic}
\usepackage{cite}
\usepackage{amsmath,amssymb,amsfonts}
\usepackage{algorithmic}
\usepackage{graphicx}
\usepackage{algorithm,algorithmic}
\usepackage{hyperref}
\hypersetup{hidelinks=true}
\usepackage{textcomp}
\usepackage{float}
\usepackage{array}
\usepackage{multirow}

\def\BibTeX{{\rm B\kern-.05em{\sc i\kern-.025em b}\kern-.08em
    T\kern-.1667em\lower.7ex\hbox{E}\kern-.125emX}}
\begin{document}
\title{Spatial and Temporal Super-Resolution and Denoising of Intracardiac 4D Flow MRI}
\author{Yusuf A. Ozturk, Jie Zheng, and Fanwei Kong
\thanks{This work was supported in part by the McKelvey CIG from Washington University, NIH R01HL192617 and NIH R01HL165238.}
\thanks{Yusuf A. Ozturk and Fanwei Kong are with the Department of Mechanical Engineering and 
Materials Science, Washington University, Saint Louis, MO, USA (e-mails: oyusuf@wustl.edu; kongf@wustl.edu).  }
\thanks{Jie Zheng is with the 
Mallinckrodt Institute of Radiology, Washington University School of Medicine, St. Louis, MO, USA (e-mail: zhengj@wustl.edu).}
}

\maketitle

\begin{abstract}
Intracardiac flow is driven by complex, and time-varying cardiac motion and interaction with the heart valves, producing complex flow features that may provide valuable information for disease diagnosis and treatment planning. 4D Flow MRI enables non-invasive, time-resolved measurement of three-dimensional cardiovascular velocity fields. However, substantial noise and limited spatial and temporal resolution limit its ability to accurately resolve flow features. We propose 4D-MRAR (4D Magnetic Resonance Augmented Reconstruction), a deep learning (DL) method to super-resolve and denoise 4D Flow MRI data. Unlike prior DL methods that train and infer on static vascular regions, our method predicts intracardiac flow within hearts undergoing large deformation. Moreover, it uniquely allows for simultaneous super-resolution in 3D space and time. The method combines a residual convolutional neural network, a recurrent neural network, and an implicit neural field to predict high-resolution velocity and pressure fields while efficiently enforcing the governing equations of fluid flow during training. Trained on a synthetic dataset obtained from computational fluid dynamics (CFD) simulations of left ventricle (LV) flow, our method generalizes to synthetic whole-heart 4D flow MRI data, and demonstrated promising performance on real 4D Flow MRI. The produced super-resolved flow fields exhibit significantly improved accuracy and greater consistency with the governing equations of blood flow compared with prior methods.
\end{abstract}


\section{Introduction}
\label{sec:introduction}
4D Flow MRI is a time-resolved, three-dimensional phase-contrast MRI technique with velocity encoding in all three spatial directions, enabling quantification of the 3D velocity fields over time in the cardiovascular system~\cite{markl20124d,stankovic20144d,sekine20224d}. The measured velocity fields can be further analyzed to derive hemodynamic metrics such as wall shear stress and kinetic energy, which can provide insights into cardiovascular disease progression and outcomes~\cite{sekine20224d,dwivedi2026alterations}. Intracardiac flow exhibits complex flow patterns such as rapidly evolving vortices, shear layers, and flow redirection driven by valvular motion and myocardial deformation. Quantitative assessments of intracardiac flow topology, vortex dynamics, and energy dissipation have been linked to the pathogenesis of various cardiovascular diseases, such as heart failure \cite{cicchitti_heart_2016}, acute myocardial infarction\cite{chan_impact_2019}, valvular heart disease\cite{pedrizzetti2014vortex}, and dilated or hypertrophic cardiomyopathy\cite{gharib_optimal_2006,pedrizzetti2014vortex}. However, these flow features occur at submillimeter spatial scales and over temporal scales  of only a few milliseconds, whereas 4D Flow MRI is limited to spatial resolutions of approximately 1.5--2.5 mm and temporal resolutions greater than 30 ms~\cite{pedrizzetti2014vortex,marchese2021left}. In addition, the signal received from the 4D Flow MRI is prone to noise and motion artifacts~\cite{peper2022advances}. Therefore, there is a pressing need to improve the spatial and temporal resolution of 4D Flow MRI while reducing noise, to enable higher-fidelity quantification of blood flow patterns and derived hemodynamic metrics for disease diagnosis and treatment planning~\cite{wu2026patient}.

Deep learning (DL) has been widely applied across medical imaging, including image segmentation, registration, reconstruction, denoising, and super-resolution~\cite{litjens_survey_2017, lundervold_overview_2019, qiu_medical_2023, montalt2021machine,liang2020deep}. Despite these advances, DL-based super-resolution of 4D Flow MRI remains relatively underexplored, particularly for resolving complex cardiovascular flow. Most existing approaches rely on convolutional neural networks (CNNs) that learn a mapping from low-resolution velocity images to high-resolution velocity fields on structured voxel grids. For example, 4DFlowNet~\cite{long2023super,ferdian2023cerebrovascular,ferdian20204dflownet} utilizes a residual CNN trained with synthetic paired low- and high-resolution data obtained from computational fluid dynamics (CFD) simulations to simultaneously denoise and spatially super-resolve 4D Flow MRI. Subsequent studies have extended this framework to neurofluid applications~\cite{patel2025super}. More recently, Callmer et al.~\cite{callmer2025deep} extended the 4DFlowNet framework by convolving in the time dimension to enable temporal super-resolution. Nonetheless, due to memory constraints, the temporal upsampling is performed on 2D slices, and the inter-slice coherency is not explicitly preserved. Patel et al.\cite{patel2026transvort} further proposed a CNN-based approach that incorporates the vorticity transport equation into the loss function. Nevertheless, these convolutional architectures produce predictions on predefined structured grids, so evaluating velocity at arbitrary spatial or temporal locations requires additional interpolation or resampling. Moreover, many among these approaches do not regularize their training with the full mass and momentum conservation equations governing blood flow, which may limit the physical fidelity of the predicted velocity fields and prevent direct recovery of pressure gradients from the learned flow solution. 

Another branch of 4D Flow MRI super-resolution methods is based on physics informed neural networks (PINNs)~\cite{raissi2019physics}. These approaches ~\cite{kalajahi2025input,fathi2020super} incorporate governing physical laws by introducing residuals of the governing equations into the training loss. PINNs typically employ coordinate-based multi-layered perceptrons (MLP) that map spatiotemporal coordinates to physical quantities,  enabling continuous prediction of the underlying fields at arbitary locations rather than restricting the predicitons to a predefined structured grid. However, PINN-based approaches are often formulated as case-specific inverse problems and therefore can be difficult to generalize across varying complex anatomies. Their convergence to the correct solutions also rely heavily on sufficient observational and boundary constraints, which are often unavailable in cardiovascular applications. Furthermore, their application to intracardiac flow with complex patient-specific geometries, moving boundaries, and highly heterogeneous flow features remains limited. 

To address these limitations, we develop a novel approach for joint spatial and temporal super-resolution of 4D Flow MRI. Our contributions are three folds. First, our approach adapted a local implicit image function (LIIF) formulation~\cite{chen2021learning} to learn a continuous spatiotemporal representation of high-resolution flow and pressure fields, enabling predictions at arbitrary spatiotemporal locations rather than on fixed structured grids.  Second, unlike prior methods focused primarily on static vascular domains, we target complex intracardiac flows within beating hearts undergoing large myocardial deformation. Third, we incorporate the full governing equations of blood flow during training, enabling the predictions of pressure gradient fields without direct pressure supervision, a capability not provided by any prior methods. Evaluations on synthetic and real 4D Flow MRI data demonstrate effective super-resolution of low-resolution inputs, with both quantitative and qualitative improvements over prior approaches. 

\section{Methods}
\subsection{Data Generation and Augmentation}
\begin{figure}[!t]
\centerline{\includegraphics[width=\columnwidth]{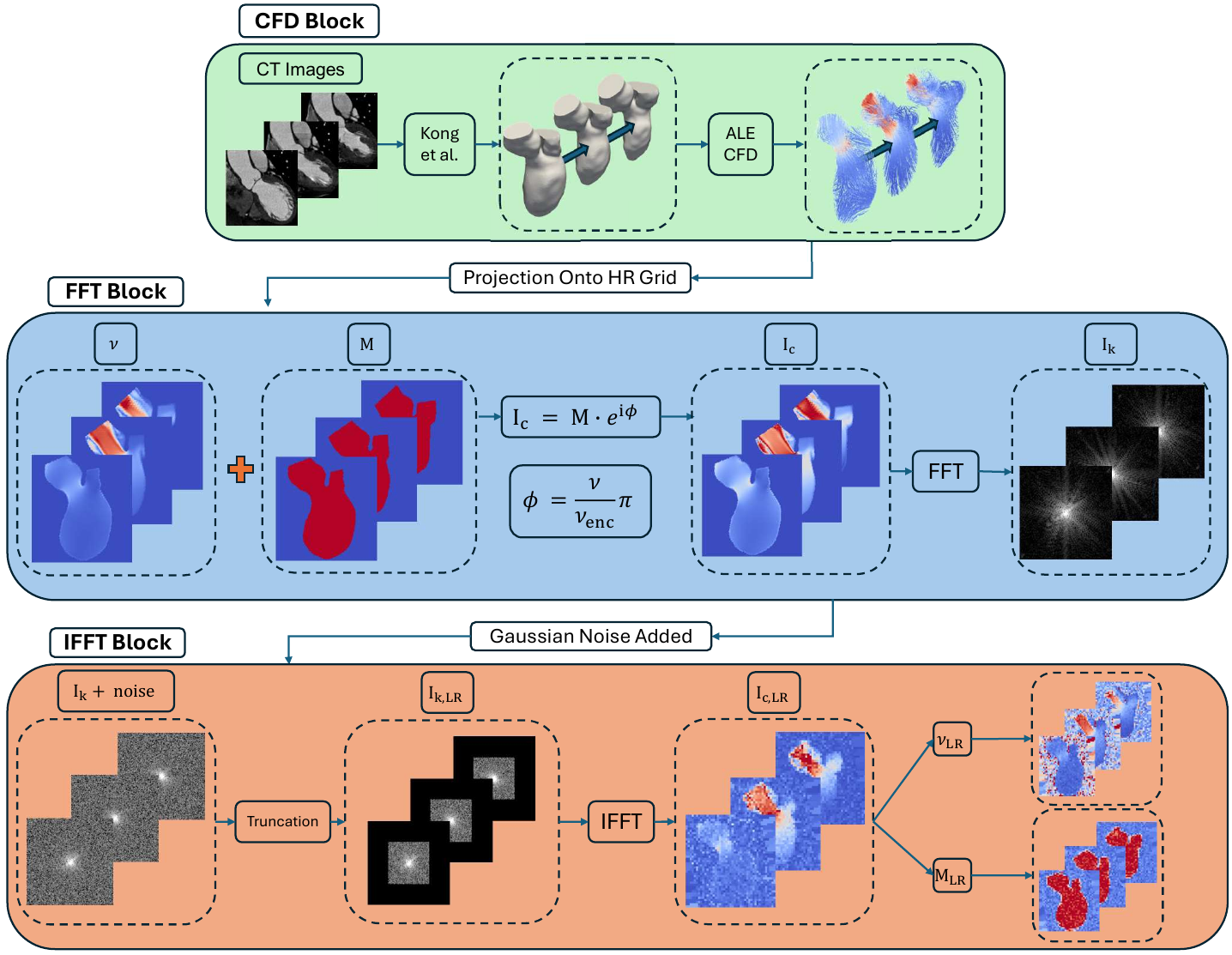}}
\caption{Synthetic 4D Flow MRI data generation pipeline. The CFD block computes synthetic velocity fields from patient-specific cardiac images. The FFT block transforms high-resolution images and synthetic magnitude images into k-space. The IFFT block adds noise and downsamples the k-space data before reconstructing low-resolution velocity and magnitude images.} \label{fig1}
\end{figure}

Our method is trained in a fully supervised manner to predict high-resolution, denoised flow fields from low-resolution input data. Due to the challenge of acquiring high-resolution in vivo 4D Flow MR data, we create synthetic paired low- and high-resolution flow fields in deforming left ventricular (LV) anatomies for training (Fig \ref{fig1}).  CFD simulations are treated as the high-resolution ground truth and the synthetic low-resolution counterparts are generated with complex and frequency space downsampling methods. 

\paragraph{Moving domain CFD} Unlike prior methods that generated synthetic training data in stationary domains,~\cite{ferdian20204dflownet,kalajahi2025input,fathi2020super}, we generate training data in dynamically deforming cardiac anatomies. Leveraging our prior DL-based framework~\cite{kong2022learning}, we reconstructed CFD-ready patient-specific deformable LV meshes from time-series CT and MRI images. We then solve the Arbitrary Lagrangian-Eulerian ~\cite{DUARTE20044819} formulation of the incompressible Navier-Stokes equation:
\begin{equation}
\rho \left.\frac{\partial u}{\partial t}\right|_{\mathbf{X}} + \rho \left[(u - u_{\text{mesh}}) \cdot \nabla\right] u + \nabla p - \mu\nabla^2 u - f = 0,
\label{eq:ale_ns}
\end{equation}
\begin{equation}
\nabla \cdot u
= 0
\label{eq:div}
\end{equation}
where $u$ is the fluid velocity, $\mathbf{X}$ is the deforming reference frame,  $u_{\text{mesh}}$ is the mesh velocity, $\rho$ is the density of the fluid, $p$ is the pressure, $\mu$ is the viscosity of the fluid,  and the $f$ is the body forces acting on the fluid. The CFD simulations are conducted with SimVascular/svFSI~\cite{Zhu2022}. Because the valves are not explicitly modeled, each valve is treated as binary open or closed depending on the cardiac phase identified from wall motion. The resulting CFD simulations provide synthetic high-resolution velocity fields on unstructured meshes.

\paragraph{k-Space Encoding} In 4D Flow MRI, velocity is encoded in the phase of complex-valued MR images and reconstructed from k-space data. To generate synthetic k-space data, we project the simulated velocity fields onto a high-resolution grid $128 \times 128 \times 128$ with an isotropic spacing of $0.625mm$ and temporal resolution of $10ms$ to obtain the high-resolution images. Since k-space encoding requires both magnitude and phase images, we use vessel masks as synthetic magnitude images $(M)$ while each velocity component is encoded and wrapped to obtain the corresponding phase image $(\phi)$ using a prescribed $v_{\mathrm{enc}}$ value. The resulting magnitude and phase images are then combined to generate a synthetic complex image $(I_c)$, which is transformed into k-space $(I_k)$ using the fast Fourier transform (FFT). 

\paragraph{Image-Space Reconstruction} To simulate measurement noise, we add Gaussian noise to the synthetic k-space images. We truncate the high-frequency components of k-space to achieve fourfold spatial downsampling. The resulting low-resolution k-space image is then transformed into complex-space using inverse FFT (IFFT), yielding complex images from which magnitude and phase are extracted; the phase is then decoded into velocity components. This produces structured low-resolution magnitude and velocity images of size $32\times 32 \times 32$ with $2.5mm$ isotropic spacing. The temporal downsampling was performed by temporal subsampling to emulate fourfold reduction in temporal resolution from $10$ ms to $40$ ms.

\paragraph{Patch Sampling} The velocity and magnitude images are divided into corresponding low- and high-resolution sub-volumes (patches). This reduces training memory and encourages learning of local flow features without relying on global anatomical cues. Temporally adjacent patch pairs are randomly sampled, with at least $20\%$ of each patch volume lies within the blood domain. 

\subsection{Network Architecture and Training} 

\begin{figure*}
\centerline{\includegraphics[width=0.85\textwidth]{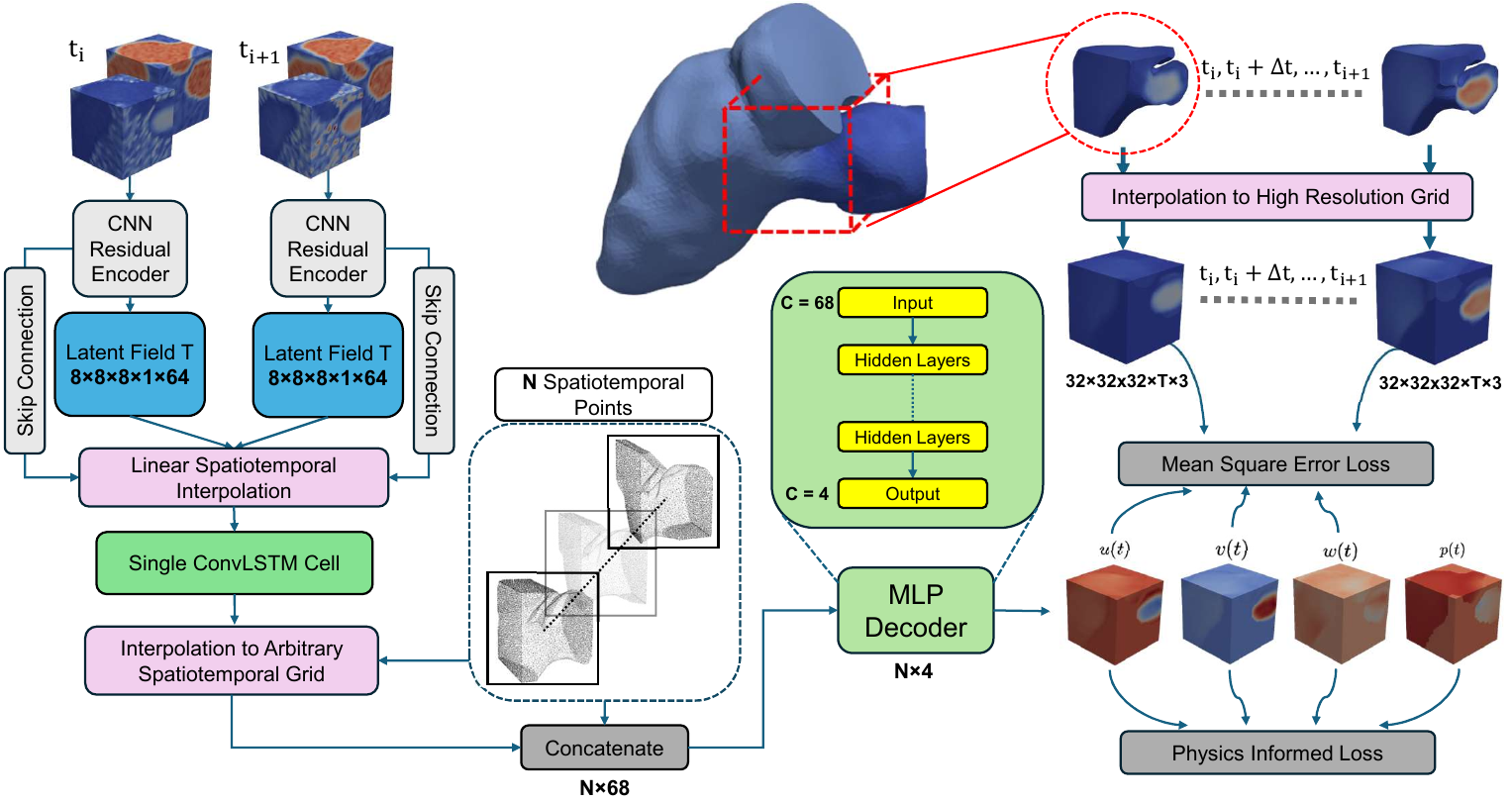}}
\caption{Overview of the 4D-MRAR architecture for spatiotemporal super-resolution of 4D Flow MRI. Low-resolution inputs are encoded into latent fields, temporally refined using a ConvLSTM module, and decoded at arbitrary spatiotemporal coordinates using a coordinate-based MLP with data and physics-informed losses.} \label{fig2}
\end{figure*}

Our framework, 4D-MRAR (Fig.~\ref{fig2}) combines the strengths of CNN and PINN to enable continuous and grid-agnostic reconstruction of flow and pressure fields at enhanced spatial and temporal resolutions with improved physical consistency. Namely, we first use a residual CNN encoder ~\cite{he2016deep,ledig2017photo} to extract a latent representation of the high-resolution flow fields from low-resolution velocity and magnitude 4D Flow MR images. We then use a temporal super-resolution module to predict the evolution of the latent field between adjacent time steps captured by the low resolution inputs. The latent representation is then passed through a coordinate-based MLP to directly query at arbitrary spatiotemporal coordinates. The mass and momentum balances are regularized during training to encourage learning of physics-consistent latent fields. 

\paragraph{Latent Field Encoding}
We train a CNN encoder to encode both the low-resolution velocity and a magnitude images into a latent representation of the underlying flow fields. The encoder has two channels: the velocity channel receives three velocity components while the magnitude channel receives the magnitude image, velocity magnitude and their element-wise product. Each channel first passes through two 3D convolution layers. The resulting features are then concatenated and processed by two additional 3D convolutional layers. Each convolutional layer is followed by a GELU~\cite{hendrycks2016gaussian} activation function, which is second-order differentiable and therefore supports the automatic differentiation required by the subsequent physics-informed loss functions. The concatenated latent features go through eight residual blocks to produce a low resolution latent field. 
To improve the spatial expressiveness of the low-resolution latent field,  we linearly upsample it in space to provide higher-resolution spatial information to the subsequent recurrent module. To reduce vanishing gradients and preserve input information, a skip connection bypasses the residual block, linking the linearly upsampled features at its input directly to the input of the subsequent recurrent module. 

\paragraph{Temporal Latent Field Super-Resolution} To learn the temporal evolution of the latent representation of high-resolution flow fields, we leverage a Convolutional Long Short Term Memory (ConvLSTM)~\cite{shi2015convolutional} module to temporally super-resolve any adjacent latent fields. Namely, the CNN encoder first independently encodes each pair of adjacent low-resolution time steps $(t_i,t_{i+1})$ into latent fields. These latent fields are first linearly upsampled in time to provide initial estimates at the target intermediate time steps. Then a ConvLSTM cell refines these estimates by introducing non-linearity and encourages temporal coherency by propagating hidden and cell states across time.

%

\paragraph{High-Resolution Flow Field Decoding} The spatiotemporally upsampled latent field is then sampled at arbitrary spatiotemporal coordinates using cubic interpolation.   The sampled latent vector at each point is then concatenated with its coordinates and fed into the MLP decoder. MLP consists of five hidden layers with the widths of $(64,48,32,32,24)$, respectively. This coordinate-based approach allows the model to query the latent field at arbitrary locations to super-resolve without being constrained to fixed grids or fixed upsampling factors unlike prior approaches~\cite{ferdian20204dflownet,callmer2025deep,patel2025super,patel2026transvort}. For instance, querying the model on a structured grid produces volumetric flow fields for conventional 4D Flow MRI visualization and analysis. Alternatively, the model can be queried directly at arbitrary coordinates, such as points on a user-defined analysis plane, enabling flow quantities to be evaluated at the desired locations without additional spatial interpolation or resampling. Each linear layer is followed by an Adaptive SiLU~\cite{ramachandran2017searching} activation. The MLP decoder outputs three velocity components and the pressure field. 


\paragraph{Physics-Informed Training} Model is trained using both physics-informed loss terms and data fidelity terms in the objective function. The data fidelity term is point-wise mean squared error (MSE) loss whereas physics-informed loss term penalizes deviations from both mass and momentum balances in blood flow. The gradients with respect to spatiotemporal coordinates of the predicted flow fields are computed using automatic differentiation. The objective function is defined as:
\begin{equation}
\mathcal{L}_{obj} = \mathcal{L}_{\mathrm{MSE}}
+
\lambda\left(
|\mathbf{M}|_2^2
+
Div^2
\right),
\end{equation}
where  $\mathcal{L}_{MSE}$, $\mathbf{M}$ and $Div$ are defined as:
\begin{equation}
\mathcal{L}_{MSE} = ||\hat{\mathbf{{v}}} - \hat{\mathbf{{u}}}||_2^2,
\end{equation}
\begin{equation}
\mathbf{M}
=
\frac{L}{UT}
\frac{\partial \hat{\mathbf{u}}}{\partial \hat{t}}
+
(\hat{\mathbf{u}}\cdot\hat{\nabla})\hat{\mathbf{u}}
-
\frac{\mu}{\rho LU}
\hat{\nabla}^{2}\hat{\mathbf{u}}
+
\hat{\nabla}\hat{p},
\end{equation}

\begin{equation}
Div = \hat{\nabla}\cdot\hat{\mathbf{u}}.
\end{equation}
where $\hat{\mathbf{{v}}}$ and $\hat{\mathbf{{u}}}$ are the normalized ground truth velocities and predicted velocities, respectively. $\mathbf{M}$ is the total momentum residual of the normalized incompressible Navier-Stokes equations and $Div$ is the mass balance equation in blood flow. Here, $L$, $U$, and $T$ denote the characteristic length, velocity, and time scales, respectively, and hatted variables represent non-dimensional quantities. During training, these losses are evaluated in an Eulerian frame, so the ALE mesh-velocity terms do not need to be considered. 
Physics-informed loss weight, $\lambda$, is selected to be $5e^{-2}$ based on hyperparameter tuning. The patch-wise training requires consistent normalization of input fields and coordinates. $L$ is invariant across patches whereas $U$ is selected as the largest magnitude of every velocity component among time adjacent input pair. $T$ is calculated as $T = L/U$ to non-dimensionalize transient terms in momentum balance. 

\section{Experiments and Results}
\subsection{Datasets}
We trained our method on synthetic 4D Flow MRI datasets generated from CFD simulations of LV blood flow at two mesh resolutions. We then evaluated the method on held-out LV anatomies from these datasets to assess generalization across unseen LV anatomies at both flow resolutions, on synthetic 4D Flow MRI generated from whole-heart CFD with valves to assess generalization to unseen anatomical regions and flow patterns, and on real 4D Flow MRI to assess generalization from synthetic training data to real measurements.
\paragraph{Coarse-resolution LV CFD} We generated a large CFD dataset across diverse LV anatomies for training and evaluation. The simulations used a maximum mesh edge size of $2$ mm and a time step of $2.5$ ms, providing sufficient resolution for flow features that are difficult to capture with 4D Flow MRI. The LV geometries and motion were generated using SDF4CHD~\cite{kong2024sdf4chd}, a generative approach to synthesize cardiac anatomies after learning a population prior from whole heart anatomies segmented from CT images. A total of $99$ simulations were generated, with $70/20/9$ cases used for training, validation, and testing, respectively. 

\paragraph{Fine-resolution LV CFD}
To evaluate the ability of our model to recover finer-scale flow features, we generated a second dataset at higher resolution. The CFD simulations used a maximum mesh edge size of $1$ mm and a time step of $1$ ms, producing flow fields with finer-scale structures that are more strongly attenuated during downsampling. $8$ simulations were generated and split into $5/1/2$ cases for training, validation, and testing, respectively. The dataset size was limited by the substantially increased computational cost due to the finer mesh and smaller time step. LV meshes and motions were obtained from 8 sets of patient gated CT images using our prior LV mesh construction framework ~\cite{kong2020automating}. A total of 4,000 and 800 patch pairs were sampled equally from the fine- and coarse-resolution training and validation datasets, respectively.

\paragraph{Whole heart CFD with valves} We tested our model on reconstructing high-resolution flow fields in a deforming whole-heart geometry with all four cardiac valves. The simulation was obtained from our previous whole-heart CFD study of a healthy subject~\cite{kong_image-based_2026}. The inclusion of valves allowed the simulation to capture realistic valve induced vortices and flow separation. The maximum mesh edge size was $2$ mm.

\paragraph{Real 4D Flow MRI} Lastly, we evaluated our performance on in vivo 4D Flow MRI acquired at clinical voxel size of $2.5 \times2.5\times2.5 \text{ mm}^{3}$ and temporal resolution of $35$ ms. Since noise-free high-resolution ground truth cannot be acquired, we evaluated the predictions qualitatively and evaluated the physical consistency quantitatively. 
The real 4D Flow MR dataset was acquired using 3T clinical scanners (Siemens MAGNETOM Prisma, Siemens Healthineers, Erlangen, Germany; Discovery MR750, GE Healthcare, Waukesha, WI, USA) in healthy young volunteers. 

\begin{figure*}[!t]
\centering

\begin{minipage}{\columnwidth}
    \centering
    \includegraphics[width=\linewidth]
    {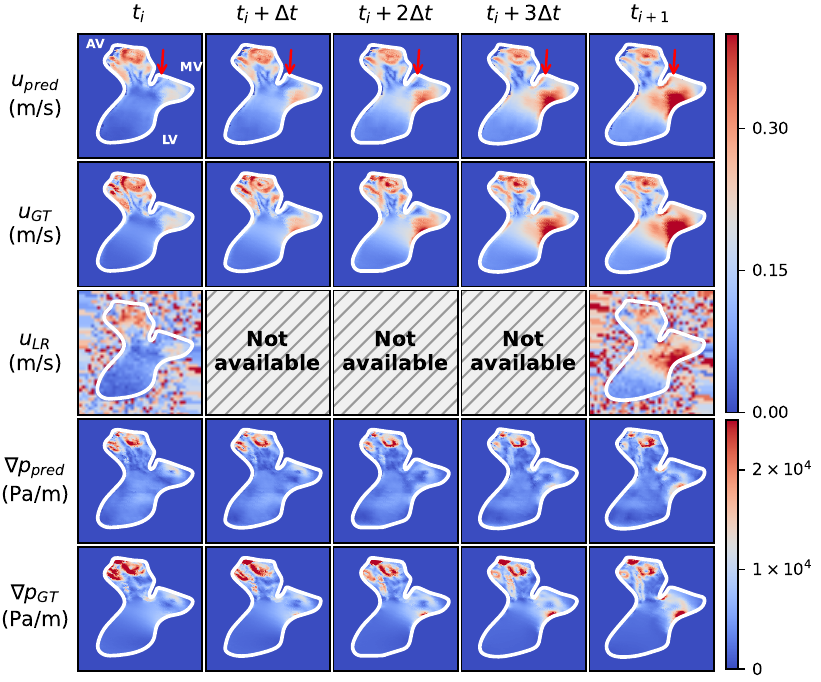}

    (a) Finer-resolution result at an early diastolic frame.
\end{minipage}
\hfill
\begin{minipage}{\columnwidth}
    \centering
    \includegraphics[width=\linewidth]
    {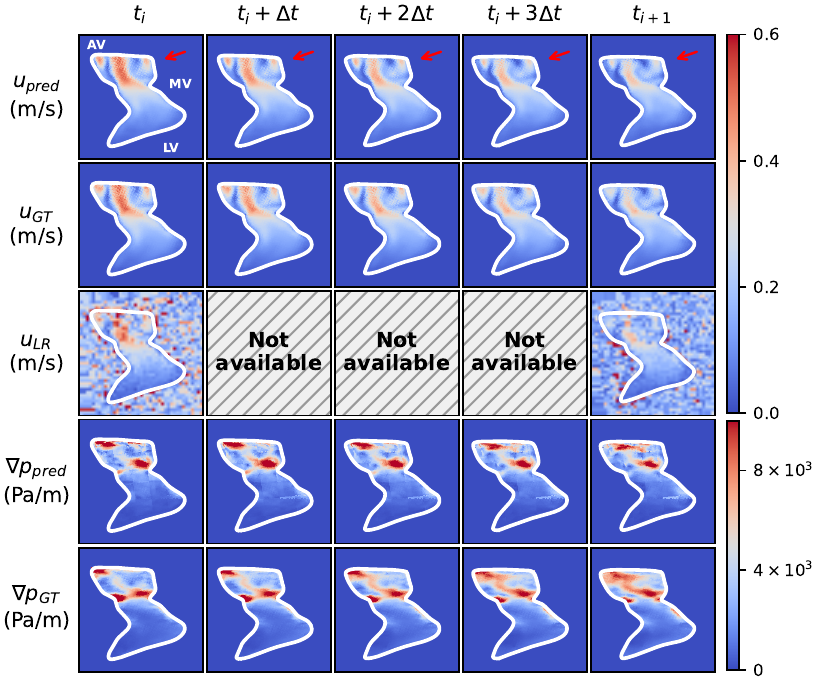}

    (b) Coarser-resolution result at a late diastolic frame.
\end{minipage}

\caption{Example predicted velocity ($u_{pred}$) and pressure gradient fields ($\nabla p_{pred}$) of our method from low-resolution inputs ($u_{LR}$) for the (a) finer-resolution CFD
and (b) coarser-resolution CFD datasets. Columns represent different
time steps ($\Delta t=10$ ms). $u_{GT}$ and $\nabla p_{GT}$ denote
the ground truth. MV and AV denote LV inlet and outlet at mitral and aortic valves, respectively.}
\label{fig4}

\end{figure*}

\subsection{Baselines and Evaluation Metrics} 
We compared our method accuracy against three baseline methods on synthetic test data and on real 4D Flow MR data: two open-source models 4DFlowNet~\cite{ferdian20204dflownet}, Callmer et al.~\cite{callmer2025deep} and an augmented PINN model adapted from Kalajahi et al.~\cite{kalajahi2025input}. Because our method performs simultaneous spatial and temporal super-resolution, whereas 4DFlowNet and Callmer et al.~\cite{callmer2025deep} address only one of these dimensions, we made the following adjustments to the training datasets to ensure compatibility with each baseline.

4DFlowNet utilizes a residual CNN with spatial convolutions, and does not perform temporal super-resolution. We therefore trained it using both edge time step (ETS) samples, where low-resolution input frames were available, and intermediate time step (ITS) where low-resolution input frames were not available. We evaluated its performance only on ETS. A recent work by Callmer et al.~\cite{callmer2025deep} uses a similar residual network but replaces 3D spatial operations with 2D+time operations. This method is designed specifically for temporal super-resolution on 2D slices and does not perform spatial super-resolution. Therefore, for this baseline, we generated a separate input dataset without spatial downsampling. Gaussian noise was added in k-space, and the reconstructed images were temporally subsampled while preserving the original spatial resolution. The model was thus trained on spatially high-resolution but temporally low-resolution input pairs and evaluated at both ETS and ITS. In Kalajahi et al. ~\cite{kalajahi2025input}, to avoid training a separate PINN for each geometry, a latent representation of low-resolution input patch is provided to the MLP together with the query point coordinates. The latent is extracted from the low-resolution input using a CNN and flattened into a global vector, in contrast to the latent field representation in our approach. 
The same scaling scheme was applied for inputs for all methods. Velocity fields were normalized by maximum magnitude among the velocity components within each patch (4DFlowNet) or patch-pair for others capable of temporal super-resolution. All models used the Adam optimizer~\cite{kingma2014adam} with an initial learning rate of $2\times10^{-5}$ and were trained until the validation loss converged.

Performance was assessed using mean absolute error, relative error, peak signal to noise ratio (PSNR),  structural similarity (SSIM), and PDE residuals ($|\mathbf{M}|$ and $Div$). Since some evaluated methods were restricted to structured grids, the PDE residuals were calculated using finite differences method. Models that lacked the ability to predict pressure fields utilized the ground truth pressure to calculate the momentum residuals $|\mathbf{M}|$. We used paired t-test for significance and 'Cohen's d'~\cite{cohen2013statistical,diener2010cohen} for effective size as statistical measures. 
Paired t-tests were performed across time steps, with predictions from different methods paired at the same time step and test case. Each case--time step combination was treated as one observation in the statistical analysis. 

\subsection{Qualitative Results on Synthetic LV 4D Flow MRI} 
We first show the performance of our method for spatial super-resolution from $2.5$ to $0.625$ mm and temporal super-resolution from $40\text{ ms}$ to $10\text{ ms}$. Fig.~\ref{fig4} demonstrates 2D slices of LV velocity and pressure gradient fields inferred from low-resolution synthetic data on test cases from the fine- and coarse- resolution CFD datasets. The proposed approach closely matched the ground truth at both ETS and ITS. In fine-resolution CFD, the ground truth flow field exhibits complex vortical flow structures that evolve over time in regions such as the aortic root. The red arrows highlight our model's ability to reasonably recover such complex flow patterns and their temporal evolution from low-resolution inputs in which the corresponding fine-scale features are largely unresolved. In coarse-resolution CFD, ground truth exhibits coarser spatial and temporal flow features. When given the corresponding lower resolution and noisy input, our model was also able to predict velocity fields that agree with ground truth. 
Our model also significantly reduced noise in both low-velocity regions, such as the apex, and high-velocity regions, such as the aortic root, compared with the noisy low-resolution inputs, demonstrating its ability to distinguish measurement noise from underlying flow features. 

In contrast to  prior approaches~\cite{ferdian20204dflownet,callmer2025deep,kalajahi2025input,fathi2020super}, our approach uniquely enabled the recovery of pressure gradient fields without pressure supervision, thanks to the momentum balance regularization. As shown in Fig. \ref{fig4}, the predicted pressure gradient fields reasonably matched the CFD ground truth. However, the pressure gradient fields exhibited more apparent patch-boundary artifacts (particularly evident near the mitral valve in Fig.\ref{fig4}). This was because the model is evaluated patch-wise and the pressure gradients are obtained from spatial derivatives of the pressure prediction, exacerbating discrepancies across patch boundaries. 



\subsection{Baseline Comparison on Synthetic LV 4D Flow MRI} 

\begin{figure}[htbp]
\centering
\includegraphics[ width=\columnwidth]{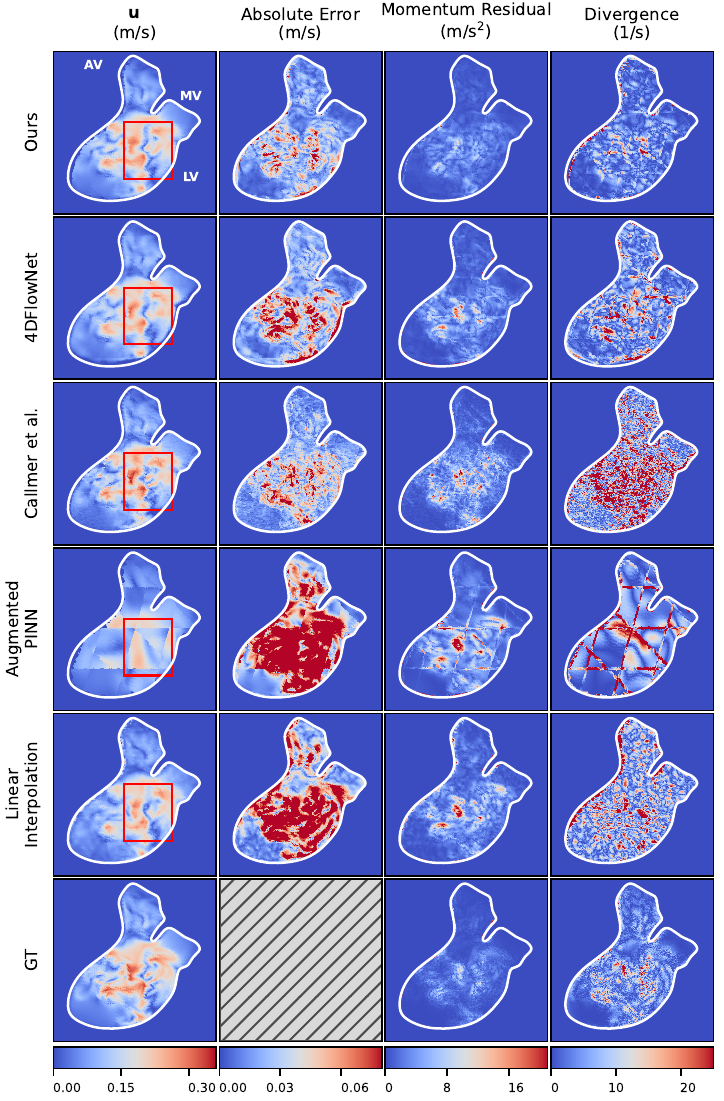}
\caption{Comparison between our model and baselines at an ETS.} \label{fig5}
\end{figure}
Fig.~\ref{fig5} demonstrates qualitative comparison between the baselines and our method on a representative late-diastolic ETS frame from the fine-resolution CFD dataset. 
Our method qualitatively outperformed the baselines in both accuracy and physical consistency, as indicated by lower absolute errors and smaller momentum and divergence residuals. Among the baselines, the augmented PINN approach was unable to reconstruct meaningful velocity fields within the moving LV domain for this test case, indicating that the per-patch global latent representation was not expressive enough to predict complex flow structures within the LV. The rest reconstructed reasonable velocity fields. However, the red boxed region indicates that 4DFlowNet and linear interpolation methods smoothed out small-scale vortical flow features. In contrast, our method and Callmer et al.~\cite{callmer2025deep} managed to preserve these features and predicted velocity fields with smaller absolute errors. Notably, our method achieved this despite using a lower-resolution input than Callmer et al. Furthermore, Callmer et al. struggled to predict a divergence free velocity field because predictions were performed independently on 2D slices, without explicitly preserving 3D spatial coherence, whereas our method and 4DFlowNet better preserved mass conservation. Likewise, since temporal coherency is not explicitly enforced in 4DFlowNet, consecutive time steps revealed inconsistent temporal fluctuations, and such inconsistent fluctuations were not observed in our method (see supplemental videos).

Tab.~\ref{comparison_table} compares the quantitative performance of our method with the baselines across all evaluation metrics on the fine- and coarse- resolution CFD datasets, across full cardiac cycles. The ITS were divided into two groups. Middle ITS (MITS) were the time steps farthest from the ETS and were expected to be the most challenging to recover. Side ITS (SITS) were the time steps adjacent to ETS. Our method significantly outperformed every baseline on all metrics for both accuracy and physics consistency. In particular, we not only significantly outperformed Callmer et al. in temporal super-resolution at both SITS and MITS, but also showed significantly better spatial super-resolution performance advantage at ETS, despite using lower-spatial-resolution inputs. 
\begin{table}[!t]
\caption{Quantitative comparison on the coarser (9 cases; 180 ETS, 360 SITS, 171 MITS patches) and finer CFD (2 cases; 30 ETS, 60 SITS, 24 MITS patches) test datasets.}
\centering
\label{comparison_table}

\resizebox{1\columnwidth}{!}{
\renewcommand{\arraystretch}{1.05}
\begin{tabular}{|c|c|c|*{4}{w{c}{1.5cm}}|}
\hline

\textbf{Dataset} & \textbf{Metric} & \textbf{Time Step}
& \textbf{Ours} & \textbf{4DFlowNet} & \textbf{Callmer et al.} & \textbf{A-PINN} \\

\hline


\multirow{18}{*}{\textbf{Coarser}}

& \multirow{3}{*}{Abs.Err (m/s)$\downarrow$}
& ETS  & \textbf{0.0028} & 0.0038$^{*\dagger}$ & 0.0042$^{*\dagger}$ & 0.0069$^{*\dagger}$ \\
& & SITS & \textbf{0.0032} & -- & 0.0046$^{*\dagger}$ & 0.0068$^{*\dagger}$ \\
& & MITS & \textbf{0.0032} & -- & 0.0045$^{*\dagger}$ & 0.0066$^{*\dagger}$ \\

\cline{2-7}

& \multirow{3}{*}{Rel.Err (\%)$\downarrow$}
& ETS  & \textbf{15.35} & 19.93$^{*\dagger}$ & 25.77$^{*\dagger}$ & 39.63$^{*\dagger}$ \\
& & SITS & \textbf{16.83} & -- & 27.53$^{*\dagger}$ & 38.32$^{*\dagger}$ \\
& & MITS & \textbf{18.07} & -- & 28.25$^{*\dagger}$ & 38.98$^{*\dagger}$ \\

\cline{2-7}

& \multirow{3}{*}{PSNR$\uparrow$}
& ETS  & \textbf{35.63} & 33.37$^{*\dagger}$ & 34.82$^{*}$ & 30.63$^{*\dagger}$ \\
& & SITS & \textbf{34.70} & -- & 34.09$^{*}$ & 30.86$^{*\dagger}$ \\
& & MITS & \textbf{34.36} & -- & 33.83$^{*}$ & 30.80$^{*\dagger}$ \\

\cline{2-7}

& \multirow{3}{*}{SSIM$\uparrow$}
& ETS  & \textbf{0.9750} & 0.9692$^{*\dagger}$ & 0.9715$^{*}$ & 0.9423$^{*\dagger}$ \\
& & SITS & \textbf{0.9705} & -- & 0.9676$^{*}$ & 0.9444$^{*\dagger}$ \\
& & MITS & \textbf{0.9691} & -- & 0.9668$^{*}$ & 0.9444$^{*\dagger}$ \\

\cline{2-7}

& \multirow{3}{*}{$M$ ($m/s^2$)$\downarrow$}
& ETS  & \textbf{0.51} & 1.61$^{*\dagger}$ & 1.82$^{*\dagger}$ & 1.71$^{*\dagger}$ \\
& & SITS & \textbf{0.52} & -- & 1.84$^{*\dagger}$ & 1.76$^{*\dagger}$ \\
& & MITS & \textbf{0.47} & -- & 1.77$^{*\dagger}$ & 1.67$^{*\dagger}$ \\

\cline{2-7}

& \multirow{3}{*}{$Div$ ($1/s$)$\downarrow$}
& ETS  & \textbf{0.87} & 1.43$^{*\dagger}$ & 2.36$^{*\dagger}$ & 1.93$^{*\dagger}$ \\
& & SITS & \textbf{0.89} & -- & 2.47$^{*\dagger}$ & 1.92$^{*\dagger}$ \\
& & MITS & \textbf{0.86} & -- & 2.40$^{*\dagger}$ & 1.84$^{*\dagger}$ \\

\hline


\multirow{18}{*}{\textbf{Finer}}

& \multirow{3}{*}{Abs.Err (m/s)$\downarrow$}
& ETS  & \textbf{0.0049} & 0.0075$^{*\dagger}$ & 0.0066$^{*\dagger}$ & 0.0126$^{*\dagger}$ \\
& & SITS & \textbf{0.0061} & -- & 0.0080$^{*\dagger}$ & 0.0129$^{*\dagger}$ \\
& & MITS & \textbf{0.0071} & -- & 0.0090$^{*\dagger}$ & 0.0136$^{*\dagger}$ \\

\cline{2-7}

& \multirow{3}{*}{Rel.Err (\%)$\downarrow$}
& ETS  & \textbf{16.16} & 24.66$^{*\dagger}$ & 21.88$^{*\dagger}$ & 42.32$^{*\dagger}$ \\
& & SITS & \textbf{19.84} & -- & 26.10$^{*\dagger}$ & 43.04$^{*\dagger}$ \\
& & MITS & \textbf{21.24} & -- & 26.71$^{*\dagger}$ & 42.18$^{*\dagger}$ \\

\cline{2-7}

& \multirow{3}{*}{PSNR$\uparrow$}
& ETS  & \textbf{35.90} & 31.15$^{*\dagger}$ & 34.91$^{*}$ & 28.31$^{*\dagger}$ \\
& & SITS & \textbf{34.23} & -- & 33.10$^{*\dagger}$ & 28.42$^{*\dagger}$ \\
& & MITS & \textbf{33.50} & -- & 32.39$^{*\dagger}$ & 28.42$^{*\dagger}$ \\

\cline{2-7}

& \multirow{3}{*}{SSIM$\uparrow$}
& ETS  & \textbf{0.9739} & 0.9448$^{*\dagger}$ & 0.9668$^{*}$ & 0.9006$^{*\dagger}$ \\
& & SITS & \textbf{0.9656} & -- & 0.9567$^{*\dagger}$ & 0.9003$^{*\dagger}$ \\
& & MITS & \textbf{0.9616} & -- & 0.9526$^{*\dagger}$ & 0.9001$^{*\dagger}$ \\

\cline{2-7}

& \multirow{3}{*}{$M$ ($m/s^2$)$\downarrow$}
& ETS  & \textbf{0.80} & 1.21$^{*\dagger}$ & 1.51$^{*\dagger}$ & 1.18$^{*\dagger}$ \\
& & SITS & \textbf{0.79} & -- & 1.48$^{*\dagger}$ & 1.24$^{*\dagger}$ \\
& & MITS & \textbf{0.86} & -- & 1.60$^{*\dagger}$ & 1.30$^{*\dagger}$ \\

\cline{2-7}

& \multirow{3}{*}{$Div$ ($1/s$)$\downarrow$}
& ETS  & \textbf{1.29} & 2.22$^{*\dagger}$ & 3.61$^{*\dagger}$ & 2.75$^{*\dagger}$ \\
& & SITS & \textbf{1.30} & -- & 3.67$^{*\dagger}$ & 2.74$^{*\dagger}$ \\
& & MITS & \textbf{1.40} & -- & 3.77$^{*\dagger}$ & 2.83$^{*\dagger}$ \\

\hline

\multicolumn{7}{|l|}{* $p<0.05$ vs. Ours (paired t-test); $\dagger$ $|d_z|\geq0.5$ vs. Ours (paired Cohen's $d_z$). \textbf{Bold} indicates best performance.}\\
\hline

\end{tabular}
}

\end{table}

\subsection{Recovering Unseen Synthetic Whole Heart Flow}

\begin{figure}
\center
\includegraphics[width=\columnwidth]{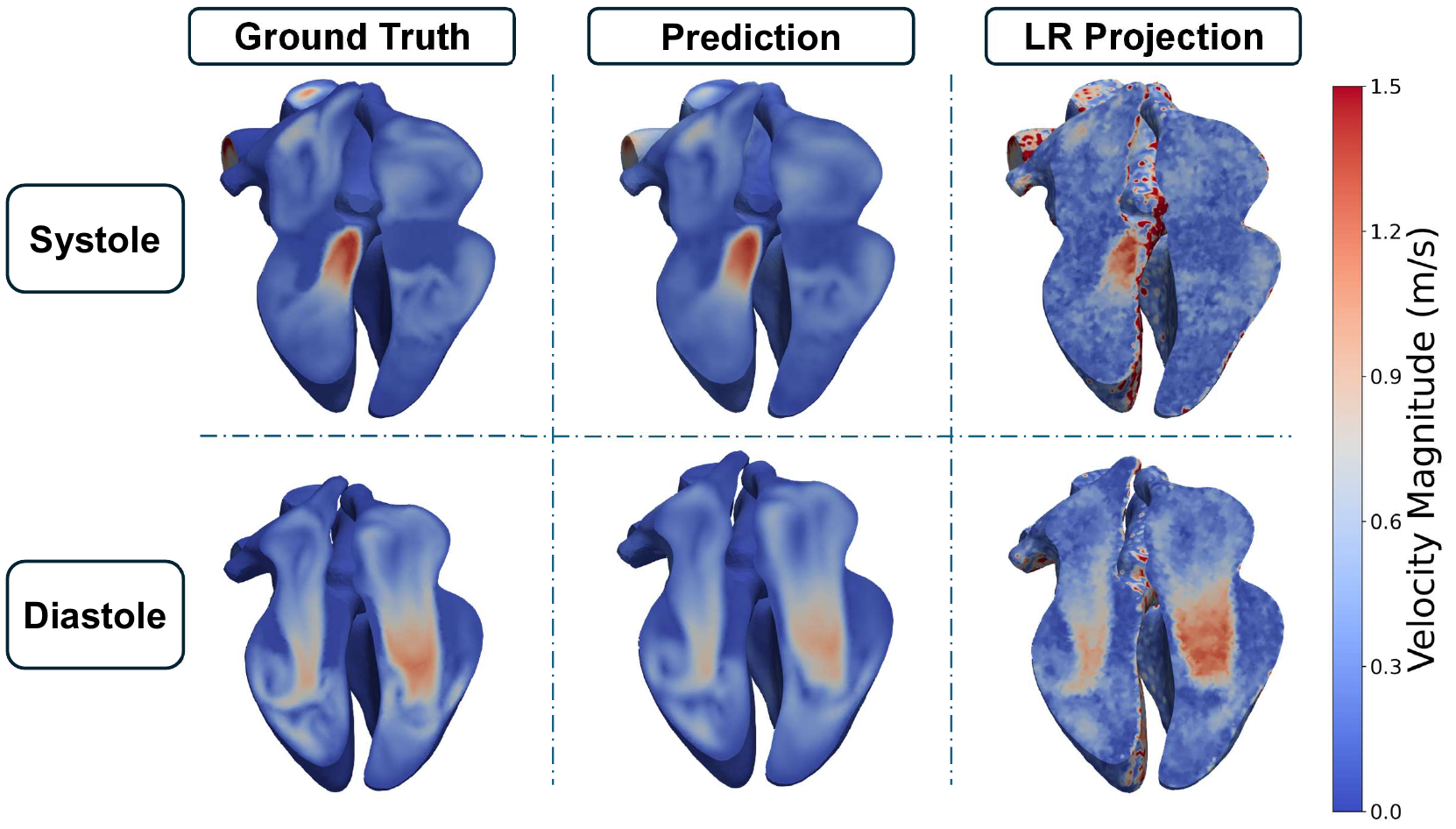}
\caption{Predicted velocity fields on a whole heart unstructured mesh from synthetic low-resolution 4D Flow MRI generated from a whole heart CFD simulation with valves at systole and diastole frames. The low-resolution input is projected onto the whole heart mesh for comparison.} 
\label{healthy_visual}
\end{figure}

By learning latent fields and decoding them with a coordinate-based continuous function represented by an MLP, our method enables predictions on unstructured meshes or arbitrary analysis planes. Fig.~\ref{healthy_visual} shows predictions on a slicing plane through a whole-heart mesh, demonstrating that the method successfully reconstructed high-resolution velocity fields in a deforming whole-heart geometry with valves despite being trained only on LV geometries without explicit valve modeling. The whole-heart CFD captured complex valve-associated flow features, including ventricular filling vortices generated by inflow through the atrioventricular valves, flow separation downstream of valve leaflets, and vortical structures in the atria and great vessels. The patch-based training, which encourages the learning of local flow features, enabled the method to generalize spatiotemporal super-resolution to these unseen cardiac structures and recover the corresponding flow patterns from low-resolution synthetic inputs.

Tab.~\ref{tablehealthy} shows a quantitative comparison between our method and the baselines on this whole-heart dataset. Our model outperformed the rest significantly in every quality metric. Our method showed some reduction in reconstruction accuracy on the unseen whole-heart dataset compared with the LV dataset used for training, but the degradation was much smaller than that observed for 4DFlowNet and Callmer et al.~\cite{callmer2025deep}. An increase in PDE residuals for all methods are observed. 

\begin{table}[!t]
\caption{Quantitative comparison on the whole heart dataset. Results were computed using 16 ETS, 31 SITS, and 15 MITS patches.}
\centering
\label{tablehealthy}
\setlength{\tabcolsep}{3pt}
\resizebox{0.8\columnwidth}{!}{
\renewcommand{\arraystretch}{1.05}
\begin{tabular}{|c|c|*{4}{w{c}{1.5cm}}|}
\hline
\textbf{Metric} & \textbf{Time Step} & Ours & 4DFlowNet & Callmer et al. & A-PINN \\
\hline
\multirow{3}{*}{Abs.Err (m/s)$\downarrow$}
& ETS  & \textbf{0.0092} & 0.0125$^{*\dagger}$ & 0.0204$^{*\dagger}$ & 0.0247$^{*\dagger}$ \\
& SITS & \textbf{0.0127} & -- & 0.0225$^{*\dagger}$ & 0.0252$^{*\dagger}$ \\
& MITS & \textbf{0.0148} & -- & 0.0236$^{*\dagger}$ & 0.0262$^{*\dagger}$ \\
\hline
\multirow{3}{*}{Rel.Err (\%)$\downarrow$}
& ETS  & \textbf{15.89} & 21.65$^{*\dagger}$ & 37.68$^{*\dagger}$ & 43.35$^{*\dagger}$ \\
& SITS & \textbf{21.56} & -- & 40.90$^{*\dagger}$ & 44.13$^{*\dagger}$ \\
& MITS & \textbf{24.63} & -- & 42.03$^{*\dagger}$ & 45.25$^{*\dagger}$ \\
\hline
\multirow{3}{*}{PSNR$\uparrow$}
& ETS  & \textbf{38.01} & 34.50$^{*\dagger}$ & 32.50$^{*\dagger}$ & 29.72$^{*\dagger}$ \\
& SITS & \textbf{35.48} & -- & 31.37$^{*\dagger}$ & 29.80$^{*\dagger}$ \\
& MITS & \textbf{34.29} & -- & 30.86$^{*\dagger}$ & 29.46$^{*\dagger}$ \\
\hline
\multirow{3}{*}{SSIM$\uparrow$}
& ETS  & \textbf{0.9696} & 0.9537$^{*\dagger}$ & 0.9248$^{*\dagger}$ & 0.8930$^{*\dagger}$ \\
& SITS & \textbf{0.9582} & -- & 0.9157$^{*\dagger}$ & 0.8895$^{*\dagger}$ \\
& MITS & \textbf{0.9521} & -- & 0.9107$^{*\dagger}$ & 0.8860$^{*\dagger}$ \\
\hline
\multirow{3}{*}{$M$ ($m/s^2$)$\downarrow$}
& ETS  & \textbf{2.89} & 3.87$^{*\dagger}$ & 5.75$^{*\dagger}$ & 3.31$^{*\dagger}$ \\
& SITS & \textbf{2.79} & -- & 5.65$^{*\dagger}$ & 3.97$^{*\dagger}$ \\
& MITS & \textbf{2.76} & -- & 5.54$^{*\dagger}$ & 3.87$^{*\dagger}$ \\
\hline
\multirow{3}{*}{$Div$ ($1/s$)$\downarrow$}
& ETS  & \textbf{2.87} & 5.12$^{*\dagger}$ & 14.17$^{*\dagger}$ & 6.01$^{*\dagger}$ \\
& SITS & \textbf{2.90} & -- & 14.18$^{*\dagger}$ & 6.67$^{*\dagger}$ \\
& MITS & \textbf{2.98} & -- & 14.08$^{*\dagger}$ & 6.68$^{*\dagger}$ \\
\hline

\multicolumn{6}{|l|}{* $p<0.05$ vs. Ours (paired t-test); $\dagger$ $|d_z|\geq0.5$ vs. Ours (paired Cohen\'s $d_z$).}\\
\multicolumn{6}{|l|}{\textbf{Bold} indicates best performance.}\\
\hline
\end{tabular}
}
\end{table}

\subsection{Baseline Comparison on Real 4D Flow MR Data} 
We further compared the methods on real 4D Flow MR data to assess their ability to generalize from synthetic training data to real acquisitions. 
\begin{figure}
\center
\includegraphics[width=\columnwidth]{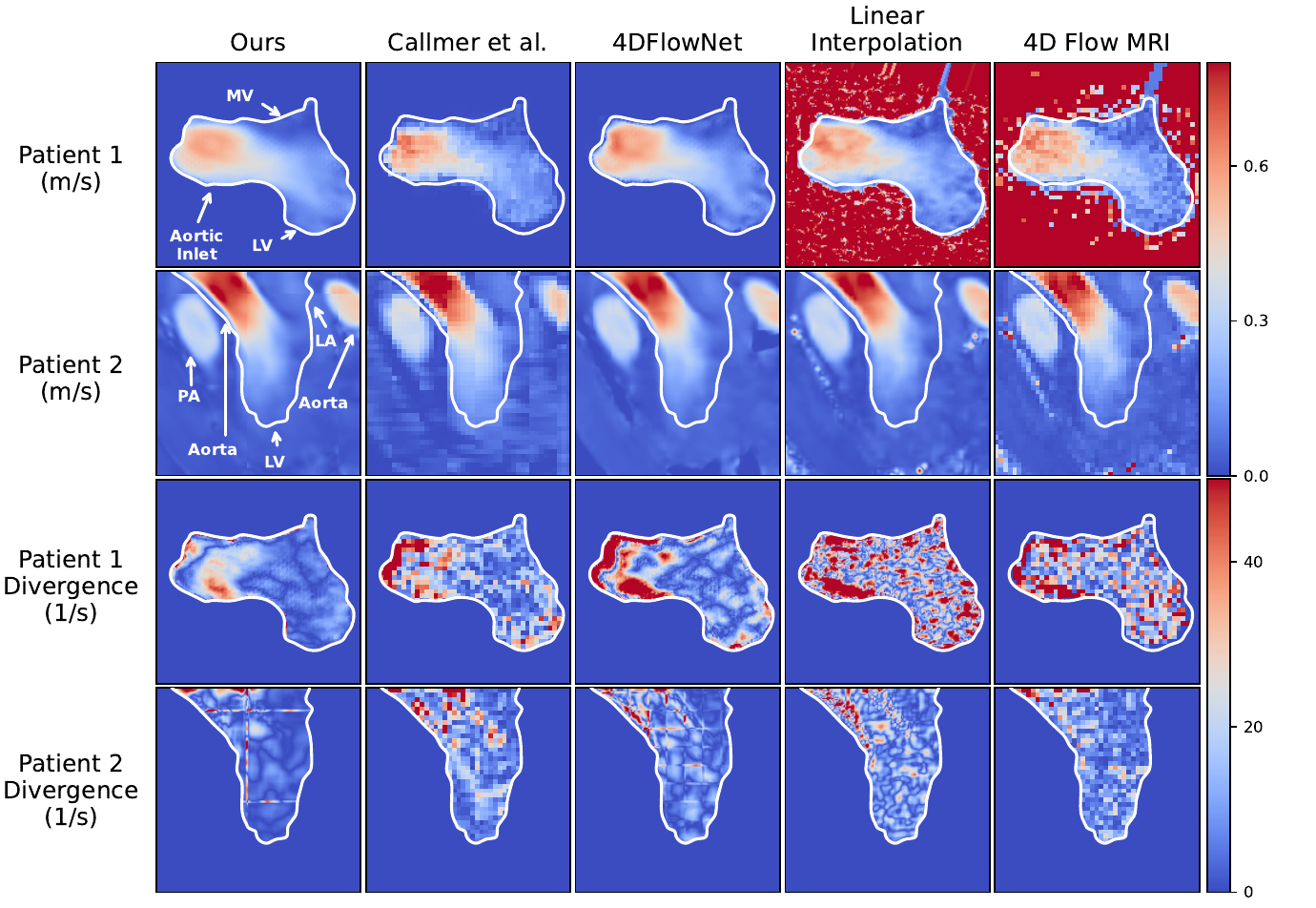}
\caption{Real 4D Flow MRI data predictions. Slices are chosen to highlight LV-aorta section of the cardiac flow. PA and LA stand for pulmonary artery and left atrium, respectively.} 
\label{fig6}
\end{figure}
Fig.~\ref{fig6} illustrates the comparison between our method and the baselines on two sets of real 4D Flow MR data. Qualitatively, our method produced less noisy velocity fields with visibly smaller divergence residuals. Callmer et al.~\cite{callmer2025deep} produced spatially low-resolution velocity fields due to the lack of spatial super-resolution capability, whereas 4DFlowNet and linear interpolation produced temporally low-resolution fields due to the lack of temporal upsampling (see supplemental videos for temporal predictions). 
For both patients, our method predicted velocity fields with the smallest divergence over a cardiac cycle compared (Tab.~\ref{table2}).

\begin{table}[!t]
\caption{Quantitative comparison study on real 4D Flow MR data -- Mean Divergence residual ($1/s$) for two patients, computed with a shared voxel spacing across all methods.}
\label{table2}
\centering

\setlength{\tabcolsep}{3pt}

\resizebox{\columnwidth}{!}{
\renewcommand{\arraystretch}{1.05}
\begin{tabular}{|c|*{5}{w{c}{1.7cm}}|}
\hline
Patient & Ours & Callmer et al. & 4DFlowNet & Linear Interp. & 4D Flow MRI \\
\hline



Patient 1 & \textbf{4.79} & 17.15 & 11.59 & 20.03 & 30.46 \\

Patient 2 & \textbf{6.59} & 16.66 & 8.34 & 15.82 & 21.11 \\

\hline
\multicolumn{6}{|l|}{\textbf{Bold} indicates best performance.}\\
\hline
\end{tabular}
}

\end{table}

\subsection{Ablation Study} 

We conducted an ablation study to evaluate the contribution of components of our method. We considered three variations of the base model: replacement of the recurrent temporal module with linear interpolation, removal of physics-based regularization, and replacement of the MLP decoder with a CNN decoder. For the CNN decoder, we applied eight consecutive residual blocks independently to each time step after spatial and temporal upsampling of the latent fields onto a high-resolution grid, followed by four convolutional branches to predict the three velocity components and pressure field. The PDE residuals were computed using finite differences, and the physics-informed loss weight $\lambda$ was tuned to $10^{-3}$.

Tab.~\ref{ablation_table} shows the ablation analysis across all evaluation metrics for both finer- and coarser-resolution CFD datasets. Since removing physics-based regularization prevents the model from predicting pressure, we utilized the ground truth pressure field to compute the momemtum residual, $M$. Removing either the recurrent temporal module or the physics-based regularization during training significantly degraded both reconstruction accuracy and physical consistency. 
\begin{table}[!t]
\caption{Quantitative comparison of ablation variants on the coarser- and finer-resolution CFD datasets.}
\centering
\label{ablation_table}

\resizebox{1\columnwidth}{!}{
\renewcommand{\arraystretch}{1.02}
\begin{tabular}{|c|c|c|*{4}{w{c}{1.5cm}}|}
\hline

\textbf{Dataset} & \textbf{Metric} & \textbf{Time Step}
& \textbf{Ours} & \textbf{Linear Interp.} & \textbf{CNN Decoder} & \textbf{No Physics} \\

\hline


\multirow{18}{*}{\textbf{Coarser}}

& \multirow{3}{*}{Abs.Err (m/s)$\downarrow$}
& ETS  & 0.0028 & 0.0032$^{*\dagger}$ & \textbf{0.0027}$^{*}$ & 0.0031$^{*\dagger}$ \\
& & SITS & \textbf{0.0032} & 0.0035$^{*\dagger}$ & 0.0035$^{*\dagger}$ & 0.0034$^{*\dagger}$ \\
& & MITS & \textbf{0.0032} & 0.0035$^{*\dagger}$ & 0.0034$^{*}$ & 0.0035$^{*\dagger}$ \\

\cline{2-7}

& \multirow{3}{*}{Rel.Err (\%)$\downarrow$}
& ETS  & \textbf{15.35} & 17.96$^{*\dagger}$ & 15.35 & 17.43$^{*\dagger}$ \\
& & SITS & \textbf{16.83} & 18.76$^{*\dagger}$ & 17.90$^{*\dagger}$ & 18.33$^{*\dagger}$ \\
& & MITS & \textbf{18.07} & 19.77$^{*\dagger}$ & 18.58$^{*}$ & 19.40$^{*\dagger}$ \\

\cline{2-7}

& \multirow{3}{*}{PSNR$\uparrow$}
& ETS  & 35.63 & 34.79$^{*\dagger}$ & \textbf{35.93}$^{*}$ & 34.70$^{*\dagger}$ \\
& & SITS & 34.70 & 34.57$^{*}$ & \textbf{34.94}$^{*}$ & 34.49$^{*}$ \\
& & MITS & 34.36 & 34.35 & \textbf{34.83}$^{*}$ & 34.25$^{*}$ \\

\cline{2-7}

& \multirow{3}{*}{SSIM$\uparrow$}
& ETS  & 0.9750 & 0.9711$^{*\dagger}$ & \textbf{0.9790}$^{*\dagger}$ & 0.9714$^{*\dagger}$ \\
& & SITS & 0.9705 & 0.9694$^{*}$ & \textbf{0.9748}$^{*\dagger}$ & 0.9697$^{*}$ \\
& & MITS & 0.9691 & 0.9689 & \textbf{0.9747}$^{*\dagger}$ & 0.9691 \\

\cline{2-7}

& \multirow{3}{*}{$M$ ($m/s^2$)$\downarrow$}
& ETS  & \textbf{0.51} & 1.03$^{*\dagger}$ & 0.72$^{*\dagger}$ & 1.65$^{*\dagger}$ \\
& & SITS & \textbf{0.52} & 0.97$^{*\dagger}$ & 0.54$^{*\dagger}$ & 1.66$^{*\dagger}$ \\
& & MITS & 0.47 & 0.88$^{*\dagger}$ & \textbf{0.45}$^{*}$ & 1.59$^{*\dagger}$ \\

\cline{2-7}

& \multirow{3}{*}{$Div$ ($1/s$)$\downarrow$}
& ETS  & 0.87 & 1.53$^{*\dagger}$ & \textbf{0.84}$^{*\dagger}$ & 1.49$^{*\dagger}$ \\
& & SITS & 0.89 & 1.49$^{*\dagger}$ & \textbf{0.70}$^{*\dagger}$ & 1.47$^{*\dagger}$ \\
& & MITS & 0.86 & 1.42$^{*\dagger}$ & \textbf{0.66}$^{*\dagger}$ & 1.40$^{*\dagger}$ \\

\hline


\multirow{18}{*}{\textbf{Finer}}

& \multirow{3}{*}{Abs.Err (m/s)$\downarrow$}
& ETS  & 0.0049 & 0.0064$^{*\dagger}$ & \textbf{0.0046}$^{*\dagger}$ & 0.0060$^{*\dagger}$ \\
& & SITS & \textbf{0.0061} & 0.0075$^{*\dagger}$ & 0.0063$^{*}$ & 0.0071$^{*\dagger}$ \\
& & MITS & \textbf{0.0071} & 0.0085$^{*\dagger}$ & 0.0072 & 0.0081$^{*\dagger}$ \\

\cline{2-7}

& \multirow{3}{*}{Rel.Err (\%)$\downarrow$}
& ETS  & 16.16 & 21.23$^{*\dagger}$ & \textbf{15.39}$^{*\dagger}$ & 19.70$^{*\dagger}$ \\
& & SITS & \textbf{19.84} & 24.70$^{*\dagger}$ & 20.20 & 23.12$^{*\dagger}$ \\
& & MITS & 21.24 & 25.74$^{*\dagger}$ & \textbf{21.15} & 24.36$^{*\dagger}$ \\

\cline{2-7}

& \multirow{3}{*}{PSNR$\uparrow$}
& ETS  & 35.90 & 32.50$^{*\dagger}$ & \textbf{36.24}$^{*}$ & 32.97$^{*\dagger}$ \\
& & SITS & \textbf{34.23} & 31.88$^{*\dagger}$ & 33.88$^{*}$ & 32.35$^{*\dagger}$ \\
& & MITS & \textbf{33.50} & 31.54$^{*\dagger}$ & 33.34 & 31.95$^{*\dagger}$ \\

\cline{2-7}

& \multirow{3}{*}{SSIM$\uparrow$}
& ETS  & 0.9739 & 0.9518$^{*\dagger}$ & \textbf{0.9753}$^{*\dagger}$ & 0.9563$^{*\dagger}$ \\
& & SITS & 0.9656 & 0.9470$^{*\dagger}$ & \textbf{0.9671}$^{*}$ & 0.9514$^{*\dagger}$ \\
& & MITS & 0.9616 & 0.9445$^{*\dagger}$ & \textbf{0.9643}$^{*}$ & 0.9487$^{*\dagger}$ \\

\cline{2-7}

& \multirow{3}{*}{$M$ ($m/s^2$)$\downarrow$}
& ETS  & \textbf{0.80} & 1.55$^{*\dagger}$ & 1.16$^{*\dagger}$ & 1.32$^{*\dagger}$ \\
& & SITS & \textbf{0.79} & 1.52$^{*\dagger}$ & 0.90$^{*\dagger}$ & 1.28$^{*\dagger}$ \\
& & MITS & \textbf{0.86} & 1.65$^{*\dagger}$ & 0.91$^{*}$ & 1.38$^{*\dagger}$ \\

\cline{2-7}

& \multirow{3}{*}{$Div$ ($1/s$)$\downarrow$}
& ETS  & 1.29 & 2.46$^{*\dagger}$ & \textbf{1.11}$^{*\dagger}$ & 2.34$^{*\dagger}$ \\
& & SITS & 1.30 & 2.28$^{*\dagger}$ & \textbf{0.95}$^{*\dagger}$ & 2.26$^{*\dagger}$ \\
& & MITS & 1.40 & 2.33$^{*\dagger}$ & \textbf{0.99}$^{*\dagger}$ & 2.38$^{*\dagger}$ \\

\hline

\multicolumn{7}{|l|}{* $p<0.05$ vs. Ours (paired t-test); $\dagger$ $|d_z|\geq0.5$ vs. Ours (paired Cohen's $d_z$). \textbf{Bold} indicates best performance.}\\
\hline

\end{tabular}
}

\end{table}
CNN decoder demonstrated significantly higher accuracy at ETS on the finer-resolution CFD dataset, whereas the difference was not significant on the coarse-resolution CFD dataset. However, the performance advantage of the CNN decoder did not extend to SITS or MITS, where the MLP decoder generally achieved lower absolute errors, with the differences being statistically significant in most comparisons. We also observed that momentum balance residuals were mostly significantly lower for MLP decoder, consistent with its more accurate prediction at ITS, where the governing equations explicitly constrain the temporal evolution of the flow fields. However, the CNN decoder exhibited significantly lower divergence residuals than the MLP decoder and demonstrated better structural similarity. The PSNR values were comparable between the CNN and MLP decoders. 

\begin{figure}
\center
\includegraphics[width=\columnwidth]{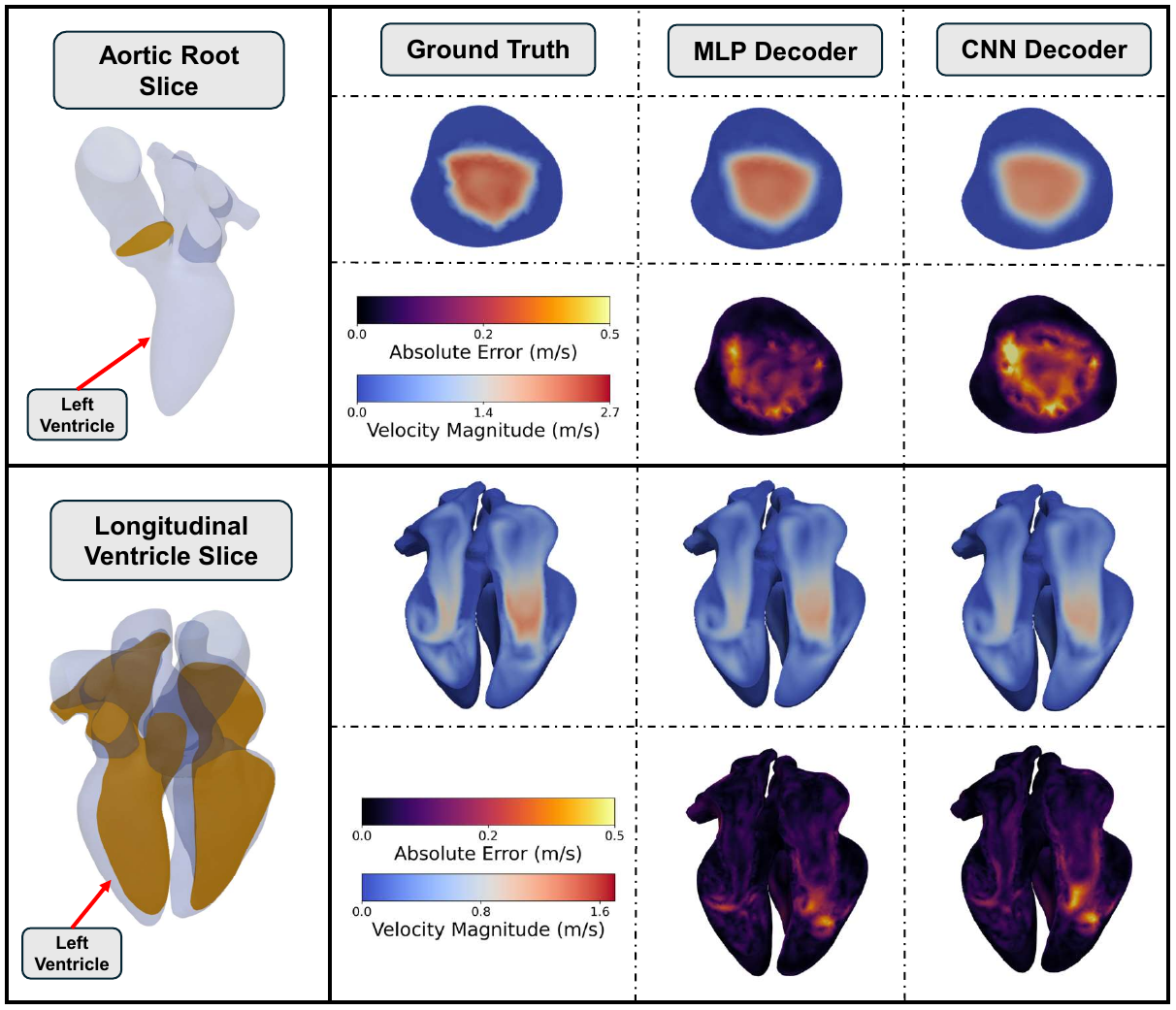}
\caption{Comparison between the coordinate-based MLP and the CNN decoder on the whole heart dataset. Velocity fields and absolute error maps are shown on an aortic-root cross-section at early systole and a longitudinal ventricular plane at early diastole.Ground-truth velocity fields are compared with MLP-decoder and trilinearly projected CNN-decoder predictions.}
\label{healthy_slicer}
\end{figure}

Although the CNN and the MLP decoders achieved, in general, comparable performance, we note that the CNN decoder reduced the model's flexibility by restricting predictions to a predefined structured grid. One of the distinguishing features of 4D-MRAR is its ability to query the learned latent field at arbitrary spatiotemporal coordinates, thereby enabling inference directly at arbitrary analysis locations or planes. Fig.~\ref{healthy_slicer} demonstrates this capability by super-resolving flow on cross-sectional slices of the aortic root and longitudinal ventricle plane, potentially supporting downstream flow-metric analysis. Because our method directly predicts flow fields at these query planes without additional interpolation, the MLP decoder generally produced velocity fields closer to the ground truth than the CNN decoder.

\section{Discussion}
4D Flow MRI is limited by noise and insufficient spatial and temporal resolution, particularly for intracardiac flow, where rapidly evolving vortices, jets, and other small-scale flow features occur within the deforming heart. Existing deep learning approaches typically address spatial or temporal super-resolution separately, operate on fixed structured grids, and have largely been developed for stationary vascular domains. Our framework addresses these problems by learning and decoding a continuous latent representation of local high-resolution flow fields while enforcing mass and momentum balances as physics-based regularization. This framework allows continuous prediction at arbitrary spatial and temporal locations and the recovery of pressure gradients without direct pressure supervision.  Across synthetic LV data, an unseen whole-heart geometry with valves, and real 4D Flow MRI, our method both qualitatively and quantitatively demonstrated improved reconstruction accuracy and physical consistency compared with existing approaches.

Notably, our method outperformed 4DFlowNet~\cite{ferdian20204dflownet} at ETS,  where both methods were given low-resolution inputs. More importantly, our method at ITS, where no low-resolution input was available, was still more accurate than 4DFlowNet at ETS for the LV CFD datasets in terms of all evaluation metrics. 
Similarly, our method outperformed Callmer et al.~\cite{callmer2025deep} in both temporal and spatial super-resolution, despite Callmer et al. receiving inputs with higher spatial resolution. Our ablation study suggests that both physics-based regularization and the ConvLSTM temporal latent evolution module contributed significantly to these improvements. Mass conservation constrains the spatial consistency of the reconstructed velocity field, while momentum balance couples its temporal evolution with spatial velocity and pressure gradients. Together with the ConvLSTM module, these constraints may encourage the latent field to encode flow features that are physically and temporally coherent, thereby helping the model distinguish true flow structures from measurement noise. As demonstrated in the Supplemental Video for both synthetic LV and \textit{in vivo} 4D Flow MRI, , this effect was particularly apparent in low-velocity regions of the LV, where the flow signal was more strongly corrupted by noise. 


Our comparison with the augmented PINN approach suggests that our performance benefits from representing local high-resolution flow fields using a spatially structured implicit latent field rather than a flattened global latent vector. Implicit representations based on spatially structured latent features have demonstrated strong expressiveness in applications such as high-resolution image reconstruction and 3D shape representation~\cite{chen2021learning, kazerouni_lift_2025}, as each latent spatial location retains localized information. Here, we extend this representation to 3D+time flow-field super-resolution. Beyond accurately reconstructing LV flow fields represented in the training data, our method generalized to unseen flow features in the whole heart, such as the valve-induced jets, flow separation, and vortical structures even though the LV training simulations did not explicitly model the valves. This suggests that 4D Flow MRI super-resolution can be learned effectively from representative local flow patterns rather than requiring knowledge of the global anatomy. This also indicates that high-fidelity representation of every anatomical feature in the training CFD used for synthetic 4D Flow MRI generation may not be necessary for training super-resolution models to recover local high-resolution flow features. 

In contrast to prior studies that have primarily focused on vascular domains with comparatively simpler pulsatile flow~\cite{fathi2020super,kalajahi2025input,patel2025super,patel2026transvort}, intracardiac flow exhibits substantially more complex and phase-dependent dynamics as the heart contracts and relaxes. Particularly in the ventricles, although the systolic flow was dominated by relatively organized high-velocity outflow, the diastolic flow exhibited more nonlinear dynamics, including vortex formation, interaction, and mixing, making temporal super-resolution particularly challenging. Indeed, when the relative errors reported in Tab.~\ref{comparison_table} were evaluated separately for systole and diastole, our method showed higher errors during diastole ($\mathord{\sim}17\%$) than systole ($\mathord{\sim}12\%$), averaged across the fine- and coarse-resolution datasets. Nevertheless, our method still significantly outperformed the baseline methods under these more challenging diastolic flow conditions, and recovered the temporal evolution of vortex structures between low-resolution input frames as illustrated in Fig. \ref{fig4}. Although clinical applications were not evaluated in this study, these capabilities may enable more detailed characterization of intracardiac vortex dynamics, which have been investigated as markers of ventricular function and cardiac disease.~\cite{toger_vortex_2012, pedrizzetti2014vortex, gharib_optimal_2006, stoll2019left}.

Although our method demonstrated promising performance in spatiotemporal super-resolution, several limitations remain to be addressed in future work to validate and improve its performance on real 4D Flow MRI. First, discrepancies remain between the real and synthetic 4D Flow MR data. 
In the current pipeline, spatial degradation is simulated primarily through truncation of high-frequency k-space components, while temporal downsampling is performed by sparsely selecting time steps. Real 4D Flow MRI acquisitions may involve more complex k-space undersampling and reconstruction schemes ~\cite{peper2022advances,stankovic20144d}, and each reconstructed cardiac phase represents measurements accumulated over a finite temporal window rather than an instantaneous flow field. Future work should better reproduce the spatial and temporal artifacts encountered in clinical 4D Flow MRI. Furthermore, high-resolution reference data are also not available for the real 4D Flow MRI, limiting quantitative validation of the reconstructed fields and leaving the clinical utility of our approach to be established. Future work will use deforming LV phantoms to acquire higher-resolution 4D Flow MRI and perform complementary validation against matched CFD simulations and particle image velocimetry (PIV)~\cite{medero2018comparison}. Lastly, although we incorporated the full governing equations of blood flow, the flow reconstruction remains under determined in the absence of explicit boundary conditions. The strong performance of our method suggests that high-resolution flow features can be reasonably inferred from the low-resolution input without requiring boundary information. Nevertheless, incorporating boundary conditions through additional boundary conditioning may further constrain the prediction, enabling the recovery of higher-resolution spatiotemporal flow features and improving robustness when the low-resolution input is uncertain, insufficient, or differs substantially from the training distribution.

\section{Conclusion}
We developed 4D-MRAR, a DL framework for simultaneous spatial and temporal super-resolution and denoising of 4D Flow MRI. By combining a CNN encoder, a recurrent temporal module, and a coordinate-based MLP decoder with physics-informed regularization, our method enables continuous reconstruction of high-resolution velocity and pressure fields at arbitrary spatiotemporal locations while improving consistency with the governing equations of blood flow. Evaluations on synthetic LV datasets demonstrated significantly improved reconstruction accuracy and physical consistency compared with existing methods, particularly in recovering complex vortical flow structures and their temporal evolution during cardiac filling. Furthermore, our method generalized to an unseen whole-heart geometry with valves, recovering valve-induced jets, flow separation, and vortical structures despite being trained only on LV simulations without explicit valve modeling. Evaluation on real 4D Flow MRI also demonstrated promising denoising performance and improved mass conservation. Overall, our results demonstrate the potential of 4D-MRAR for resolving complex intracardiac flow dynamics from 4D Flow MRI.


\section*{References}
\bibliographystyle{IEEEtran}
\bibliography{mylib}

\end{document}